\newcommand{\paperTitle}{\texttt{TranCIT}: Transient Causal Interaction Toolbox}

\newcommand{\authorone}{Salar Nouri}
\newcommand{\authortwo}{Kaidi Shao}
\newcommand{\authorthree}{Shervin Safavi}
\newcommand{\afillA}{School of Electrical and Computer Engineering, College of Engineering, University of Tehran, Tehran, Iran}
\newcommand{\afillB}{International Center for Primate Brain Research (ICPBR), Center for Excellence in Brain Science and Intelligence Technology (CEBSIT), Chinese Academy of Sciences (CAS), Shanghai, China}
\newcommand{\afillC}{Computational Neuroscience, Department of Child and Adolescent Psychiatry, Faculty of Medicine, Technische Universität Dresden, Dresden 01307, Germany}
\newcommand{\afillD}{Department of Computational Neuroscience, Max Planck Institute for Biological Cybernetics, Tübingen 72076, Germany}

\documentclass[9pt,biorxiv]{lapreprint}

\usepackage[version=4]{mhchem} 
\usepackage{siunitx}    
\usepackage{pdflscape}  
\usepackage{rotating}   
\usepackage{textgreek}  
\usepackage{gensymb}    
\usepackage[misc]{ifsym} 
\usepackage{orcidlink}  
\usepackage{listings}   
\usepackage{colortbl}   
\usepackage{tabularx}   
\usepackage{longtable}  
\usepackage{subcaption}
\usepackage{multirow}
\usepackage{snotez}     
\usepackage{csquotes}   
\usepackage{svg}
\usepackage[most]{tcolorbox}
\usepackage{afterpage}

\DeclareSIUnit\Molar{M}

\title{\paperTitle}

\author[ \orcidlink{0000-0002-8846-9318} 1 \Letter]{\authorone}
\author[ \orcidlink{0000-0002-3027-0090} 2 \Letter]{\authortwo}
\author[ \orcidlink{0000-0002-2868-530X} 3, 4 \Letter]{\authorthree}

\affil[1]{\afillA}
\affil[2]{\afillB}
\affil[3]{\afillC}
\affil[4]{\afillD}
\correspondence{salar.nouri@ut.ac.ir; kaidi.shao@icpbr.ac.cn; research@shervinsafavi.org; }{SN,  KS, SS}

\leadauthor{Nouri}
\shorttitle{\texttt{TranCIT}: Transient Causal Interaction Toolbox}

\usepackage[numbers,compress,sort,square,comma]{natbib}
\usepackage{xfrac}

\usepackage{xspace}

\usepackage[textsize = tiny,backgroundcolor=blue!8!white]{todonotes}
\newcommand{\cut}[1]{}

\newcommand{\shsc}[1]{}
\newcommand{\shsd}[1]{}
\newcommand{\shstd}[1]{}

\newcommand{\olishd}[1]{} 

\begin{document}
\maketitle



\begin{abstract}
    Quantifying transient causal interactions from non-stationary neural signals is a fundamental challenge in neuroscience. Traditional methods are often inadequate for brief neural events, and advanced, event-specific techniques have lacked accessible implementations within the Python ecosystem. Here, we introduce \texttt{trancit} (Transient Causal Interaction Toolbox), an open-source Python package designed to bridge this gap. TranCIT implements a comprehensive analysis pipeline, including Granger Causality, Transfer Entropy, and the more robust Structural Causal Model-based Dynamic Causal Strength (DCS) and relative Dynamic Causal Strength (rDCS) for accurately detecting event-driven causal effects. We demonstrate TranCIT's utility by successfully capturing causality in high-synchrony regimes where traditional methods fail and by identifying the known transient information flow from hippocampal CA3 to CA1 during sharp-wave ripple events in real-world data. The package offers a user-friendly, validated solution for investigating the transient causal dynamics that govern complex systems.
\end{abstract}


\section{Summary}

Understanding \emph{transient} causal interactions in the brain (e.g., between neurons or neural populations) based on neural signals during neural events (transient and characteristic neural activities, e.g., Sharp Wave-Ripples, see \cite{jooHippocampalSharpWave2018}) is essential for understanding the large-scale neural dynamics underlying numerous cognitive functions \citep{womelsdorfBurstFiringSynchronizes2014, lundqvistBetaBurstsCognition2024,safavi2022multistability,dwarakanath2023bistability}. Furthermore, such interactions can happen across a range of spatial scales, from neurons and neural populations \citep{liuECannulaRevealsAnatomical2022b, safaviUncoveringOrganizationNeural2023, safaviUnivariateMultivariateCoupling2021, rasch2008InferringSpikeTrains, raschNeuronsCircuitsLinear2009} to large-scale inter-area interactions \citep{logothetisHippocampalCorticalInteraction2012, nitzanBrainwideInteractionsHippocampal2022}, which further highlights their importance. Traditional causality estimation methods, such as Granger causality \cite{granger1969investigating} and Transfer Entropy \citep[TE, ][]{Schreiber2000}, often assume stationarity and require long data segments, making them suboptimal for the short-duration, non-stationary transient events frequently observed in neural data \cite{mitra2007observed}.

We present \texttt{trancit}, a robust, open-source Python package that implements advanced causal inference methods designed explicitly for transient neural events \citep[originally developed by][]{shao2023transient}. TranCIT toolkit facilitates dynamic causal inference on time series data derived from neuroscience experiments, such as electrophysiology data, electroencephalography (EEG), or magnetoencephalography (MEG). This package reimplements and extends the software of the dynamic causal learning algorithm originally introduced in MATLAB by \citet{shao2023transient}. Here we adapt the original MATLAB software to the widely used Python ecosystem. By leveraging foundational Python libraries such as NumPy \citep{harris2020array}, SciPy \citep{virtanen2020fundamental}, and Matplotlib \citep{hunter2007matplotlib}. Thus, the TranCIT toolkit enables integration with existing data science and neuroscience workflows, making it accessible to a broader research community, particularly in neuroscience and machine learning.
The package also offers an integrated solution for causal analysis, including:
\begin{itemize}
    \item \textbf{Causal inference methods:} Algorithms to detect and quantify causal relationships, including Granger Causality (GC), Transfer Entropy (TE), Dynamic Causal Strength (DCS), and relative Dynamic Causal Strength (rDCS). These methods are tailored for analyzing potentially time-varying influences that are crucial for studying dynamic brain processes.
    
    \item \textbf{Event-based preprocessing:} A limited set of routines to detect transient event timings, align epochs relative to these events (e.g., \textbf{local} signal peaks), extract relevant data snapshots including necessary time lags, and optionally reject artifact-contaminated epochs based on amplitude thresholds.
    
    \item \textbf{Simulation tools:} Utilities to generate synthetic autoregressive (AR) time series data with predefined causal structures, enabling users to validate methods, test hypotheses, or explore theoretical scenarios in a controlled environment.
\end{itemize}

\section{Statement of need}

Understanding causal interactions in the brain is a cornerstone of neuroscience research \cite{Seth2015, weichwald2021causality, ross2024causation}, shedding light on how neural populations influence each other and coordinate during cognitive tasks (e.g., memory encoding, decision-making), and sensory processing.
While widely used statistical methods often focus on correlation, causality offers deeper insight into directional influences and, most importantly, provides opportunities to gain a mechanistic understanding of the underlying relationships.
In particular, here we focused on causal inference for transient dynamics. The original dynamic causal learning algorithm by \citet{shao2023transient} addressed the need for analyzing transient events by investigating causal interactions; however, it was implemented in MATLAB, which may limit accessibility.

To bridge this gap, we introduce \texttt{trancit}, an open-source Python-based implementation, aligning with the common use of Python in neuroscience \cite{muller2015python} and data science. Notably, established libraries like \texttt{MNE-Python} (for EEG/MEG analysis) \cite{MNEPython2013}, \texttt{Nilearn} (for fMRI), and \texttt{Brainstorm} \cite{Brainstorm2011} have solidified Python's role in the field. However, existing general-purpose Python tools for causal inference, such as \texttt{causal-learn} \cite{zheng2024causal} or \texttt{tigramite} \cite{runge2022jakobrunge}, often lack specialized features optimized for the transient characteristics of neural data, like handling transient events or integrating easily with event-based preprocessing workflows.

TranCIT toolkit addresses these specific needs by offering a tailored solution that implements GC, TE, DCS, and rDCS with configurations suitable for potentially non-stationary, event-related brain data, alongside simulation utilities designed to mimic such signals. By making these capabilities freely available, \texttt{trancit} promotes reproducible research, lowers barriers to entry for advanced causal inference in neuroscience, and supports a wide range of applications.
\section{Functionality}

\texttt{trancit} package provides a comprehensive workflow for causal inference, with a primary focus on preprocessing relevant to transient events, core causal analysis methods, and simulation capabilities.

\subsection{Preprocessing}

Preparing neural time series data for causal analysis is crucial for drawing reliable conclusions. The TranCIT provides an integrated workflow focused on identifying and extracting relevant transient events from the data before causal estimation:

\begin{itemize}
    \item \textbf{Event detection and alignment}: The process begins by identifying candidate event timings through thresholding applied to a designated detection signal. Users can configure the sensitivity via parameters like a threshold ratio relative to signal variance. Once candidate points are identified, the package offers configurable strategies to align the timing of each event precisely (it can also be expanded with more sophisticated methods \cite{einevollModellingAnalysisLocal2013, safaviBrainComplexSystem2022}). Alignment can be set to capture the local signal peak near the detected point or employ a pooling method across nearby detection points, optionally combined with uniform resampling techniques to refine the event set.

    \item \textbf{Epoch extraction (snapshotting)}: Following alignment, the software extracts fixed-length data segments, or "snapshots," centered around each validated event time point. The duration of these epochs and their starting offset relative to the alignment point are user-defined parameters. This step automatically includes the necessary historical (lagged) data within each snapshot, corresponding to the model order specified for the subsequent causality analysis.

    \item \textbf{Artifact rejection}: To enhance data quality, an optional step that allows for the exclusion of entire event epochs (trials) that may be contaminated by large artifacts. This rejection is based on whether the signal amplitude within an epoch crosses a specified threshold, effectively removing trials with potentially non-physiological signal excursions.
\end{itemize}

\noindent\textit{(Note: Standard preprocessing like filtering or normalization should be applied before using \texttt{trancit}, utilizing libraries like SciPy or MNE-Python.)}

These integrated steps ensure that the subsequent causal analysis is performed on well-defined, event-aligned data segments that are relatively free from major artifacts, thereby improving the reliability of the inferred causal relationships.

\subsection{Causal inference methods}

The package implements four primary methods to detect and quantify causal relationships:

\begin{itemize}
    \item \textbf{Granger Causality (GC)} \cite{granger1969investigating}: This approach assesses whether historical data from one time series can enhance the prediction accuracy of another time series beyond using its past alone. It employs Vector Autoregressive (VAR) models and is particularly useful for determining linear predictive relationships between variables.

    The \texttt{trancit} library supports multivariate Granger causality testing and allows the adjustment of model order (lags). Specifically, GC evaluates causality by comparing predictions from two autoregressive models:

    \begin{itemize}
        \item \textbf{Univariate (reduced) model:}
        $Y_t = \sum_{i=1}^{p} a_i Y_{t-i} + \epsilon_t$

        \item \textbf{Bivariate (full) model:}
        $Y_t = \sum_{i=1}^{p} a_i Y_{t-i} + \sum_{i=1}^{p} b_i X_{t-i} + \eta_t$
    \end{itemize}

    If incorporating past values of $X$ significantly reduces the prediction error (typically assessed through an F-test), we conclude that $X$ Granger causes $Y$.

    Quantitatively, Granger causality is computed as the logarithmic ratio of the residual variances between the reduced and full models:
    \begin{equation}
    \begin{aligned}
        GC(X_t^2 \rightarrow X_t^1) = \frac{1}{2} \log \left(\frac{\hat{\sigma}'^2_{1,t}}{\hat{\sigma}^2_{1,t}}\right)\,.
    \end{aligned}
    \label{eq: granger-causality}
    \end{equation}
    Here, $\hat{\sigma}'^2_{1,t}$ represents the estimated residual variance from the reduced model (predicting $X^1_t$ using only its past), and $\hat{\sigma}^2_{1,t}$ denotes the residual variance from the full model (predicting $X^1_t$ using the past values of both $X^1$ and $X^2$) \cite{shao2023transient}.

    \item \textbf{Transfer Entropy (TE)} \cite{Schreiber2000}: It quantifies the reduction in uncertainty regarding the future state of a target time series \(Y\) when incorporating historical information from source series \(X\), given the past of \(Y\) itself. It effectively captures directed information flow, including both linear and non-linear interactions. The \texttt{trancit} library enables customization of the length of historical dependencies (time lags).

    Formally, TE is defined in terms of conditional mutual information:
    \begin{equation}
    \begin{aligned}
        TE(X \rightarrow Y) = I(Y_t ; X_{t-1} | Y_{t-1})\,.
    \end{aligned}
    \label{eq: transfer-entropy-mi}
    \end{equation}
    TE can be equivalent to the difference between conditional entropies or the expectation of the Kullback–Leibler (KL) divergence between conditional probability distributions.

    In linear, time-inhomogeneous Structural Vector Autoregressive (SVAR) models, TE compares the conditional distribution of $X_{1,t}$ given its past and the past of $X_2$ against the distribution conditioned only on the past of $X_1$. Specifically, for Gaussian SVAR models, TE has the analytical expression:
    \begin{equation}
    \begin{aligned}
        TE(X_{2,t} \rightarrow X_{1,t}) = \frac{1}{2}\log\left(\frac{\sigma_{1,t}^2 + b_t^\top \operatorname{Cov}[X_{2p,t}|X_{1p,t}]b_t}{\sigma_{1,t}^2}\right)\,.
    \end{aligned}
    \label{eq: transfer-entropy}
    \end{equation}
    Here:
    \begin{itemize}
        \item $\sigma_{1,t}^2$ is the innovation variance of the full SVAR model.
        \item $b_t$ are SVAR coefficients that quantify the influence of the past values of $X_2$ on the current value of $X_1$.
        \item $\operatorname{Cov}[X_{2p,t}|X_{1p,t}]$ represents the conditional covariance matrix of the past values of $X_2$ given the past values of $X_1$.
    \end{itemize}

    \item \textbf{Dynamic Causal Strength (DCS)} \cite{shao2023transient} DCS is a time-dependent adaptation of the Causal Strength (CS) \cite{janzing2013quantifying} measure, tailored explicitly for time-inhomogeneous SVAR models and their associated causal graphs. Similar to CS, DCS employs the Structural Causal Model (SCM) interventional formalism to quantify the magnitude of causal influence, enabling the assessment of how causal relationships change dynamically over time. DCS quantifies causal influence by measuring the KL divergence between two conditional probability distributions:
    \begin{enumerate}
        \item The original conditional distribution of the effect variable $X_{1,t}$, given its causal parents $X_{1p,t}, X_{2p,t}$.
        \item The conditional distribution following an intervention designed to eliminate the causal influence from $X_{2p,t}$ to $X_{1,t}$. This intervention replaces $X_{2p,t}$ in the structural equation for $X_{1,t}$ with an independent copy $X_{2p,t}'$, sampled from the current marginal distribution $p(X_{2p,t})$, thus removing direct causal dependence. Dependencies across successive time points within $X_2$ are preserved.
    \end{enumerate}
    The formal definition of DCS is:
    \begin{equation}
    \begin{aligned}
    DCS(X_{2,t} \rightarrow X_{1,t}) &= \mathbb{E}_{X_{1p,t}, X_{2p,t}} \Big[ D_{KL} \big( p(X_{1,t}|X_{1p,t},X_{2p,t}) \\
    &\quad \parallel\ p_{do(X_{1,t}:=f(X_{1p,t},X_{2p,t}',\eta_{1,t}))}(X_{1,t}|X_{1p,t},X_{2p,t}) \big) \Big]\,,
    \end{aligned}
    \label{eq: dcs}
    \end{equation}
    where $X_{2p,t}'$ is independently drawn from the marginal distribution $p(X_{2p,t})$.

    For Gaussian SVAR models, DCS can be expressed analytically as:
    \begin{equation}
    \begin{aligned}
      DCS(X_{2,t} \rightarrow X_{1,t}) = \frac{1}{2} \log\left(\frac{b_t^\top \operatorname{Cov}[X_{2p,t}] b_t + \sigma_{1,t}^2}{\sigma_{1,t}^2}\right)\,.
    \end{aligned}
    \label{eq: dcs-gaussian-svar}
    \end{equation}
    In this expression:
    \begin{itemize}
        \item $\sigma_{1,t}^2$ is the innovation variance of the variable $X_{1,t}$.
        \item $b_t$ represents the SVAR coefficients that link the past of $X_2$ to the current value of $X_1$.
        \item $\operatorname{Cov}[X_{2p,t}]$ is the marginal covariance matrix of the past values of $X_2$.
    \end{itemize}

    \item \textbf{Relative Dynamic Causal Strength (rDCS)} \cite{shao2023transient}: rDCS is developed as a variation of DCS, specifically tailored to assess event-based causality. It addresses situations where deterministic perturbations occur in the causal variable by quantifying causal effects relative to a baseline or reference state. Similar to DCS, rDCS utilizes interventions within the SCM framework, quantified as the expected KL divergence between two conditional distributions. However, the intervention for rDCS differs from that of DCS:
    \begin{itemize}
        \item Rather than substituting the causal variable’s past input with a sample from its current marginal distribution, rDCS substitutes it with a sample drawn from the marginal distribution at a reference baseline time point $t_{ref}$. The reference period $t_{ref}$ is typically selected during a stationary or baseline interval preceding the transient event.
    \end{itemize}
    The general definition of rDCS is:
    \begin{equation}
    \begin{aligned}
    rDCS(X_{2,t} \rightarrow X_{1,t}) = \mathbb{E}_{X_{1p,t}, X_{2p,t}} \Big[
        D_{KL}\Big( &p(X_{1,t} \mid X_{1p,t}, X_{2p,t}) \; \| \\
        &p_{do(X_{1,t} := f(X_{1p,t}, X_{2p,t_{ref}}, \eta_{1,t}))}(X_{1,t} \mid X_{1p,t}, X_{2p,t}) \Big)\,,
    \Big]
    \end{aligned}
    \label{eq: rdcs}
    \end{equation}
    where $X_{2p,t_{ref}}$ is sampled from the marginal distribution $p_{ref}(X_{2p,t_{ref}})$ at the baseline reference time.

    For Gaussian SVAR models, rDCS is analytically expressed as:
    \begin{equation}
    \begin{aligned}
    rDCS(X_{2,t} \rightarrow X_{1,t}) = \; &\frac{1}{2} \log \left(
    \frac{\sigma_{1,t}^2 + b_t^\top \operatorname{Cov}[X_{2p,t_{ref}}] b_t}{\sigma_{1,t}^2}
    \right) - \frac{1}{2} \\
    &+ \frac{1}{2} \cdot \frac{b_t^\top \mathbb{E}\left[
    (X_{2p,t} - \mathbb{E}[X_{2p,t_{ref}}])(X_{2p,t} - \mathbb{E}[X_{2p,t_{ref}}])^\top
    \right] b_t}{\sigma_{1,t}^2 + b_t^\top \operatorname{Cov}[X_{2p,t_{ref}}] b_t}\,.
    \end{aligned}
    \label{eq: rdcs-gaussian-svar}
    \end{equation}
    In this expression:
    \begin{itemize}
        \item $\sigma_{1,t}^2$ denotes the innovation variance of $X_{1,t}$.
        \item $b_t$ are SVAR coefficients quantifying the influence from past values of $X_2$ to the present $X_1$.
        \item $\operatorname{Cov}[X_{2p,t_{ref}}]$ represents the covariance matrix of $X_2$'s past values at the baseline reference state.
        \item $\mathbb{E}\left[(X_{2p,t}-\mathbb{E}[X_{2p,t_{ref}}])(X_{2p,t}-\mathbb{E}[X_{2p,t_{ref}}])^\top\right]$ captures the expected outer product of the deviation of the current state from the mean of the baseline reference state.
    \end{itemize}
\end{itemize}

\subsection{Interventions and the interventional distribution}

\noindent\textbf{Why this matters?} 
Our causal measures (GC, TE, DCS, rDCS) are most straightforward to interpret once readers understand interventions as 'what would happen if we removed a causal link or clamped an input.' 
In SCMs, interventions replace the structural assignment of a node and thereby \textbf{cut the arrows from its parents}; the resulting post-manipulation (''do-'') distribution is the quantitative object we compare to the observational one. 

\begin{figure}[t]
    \centering
    \includegraphics[width=\textwidth]{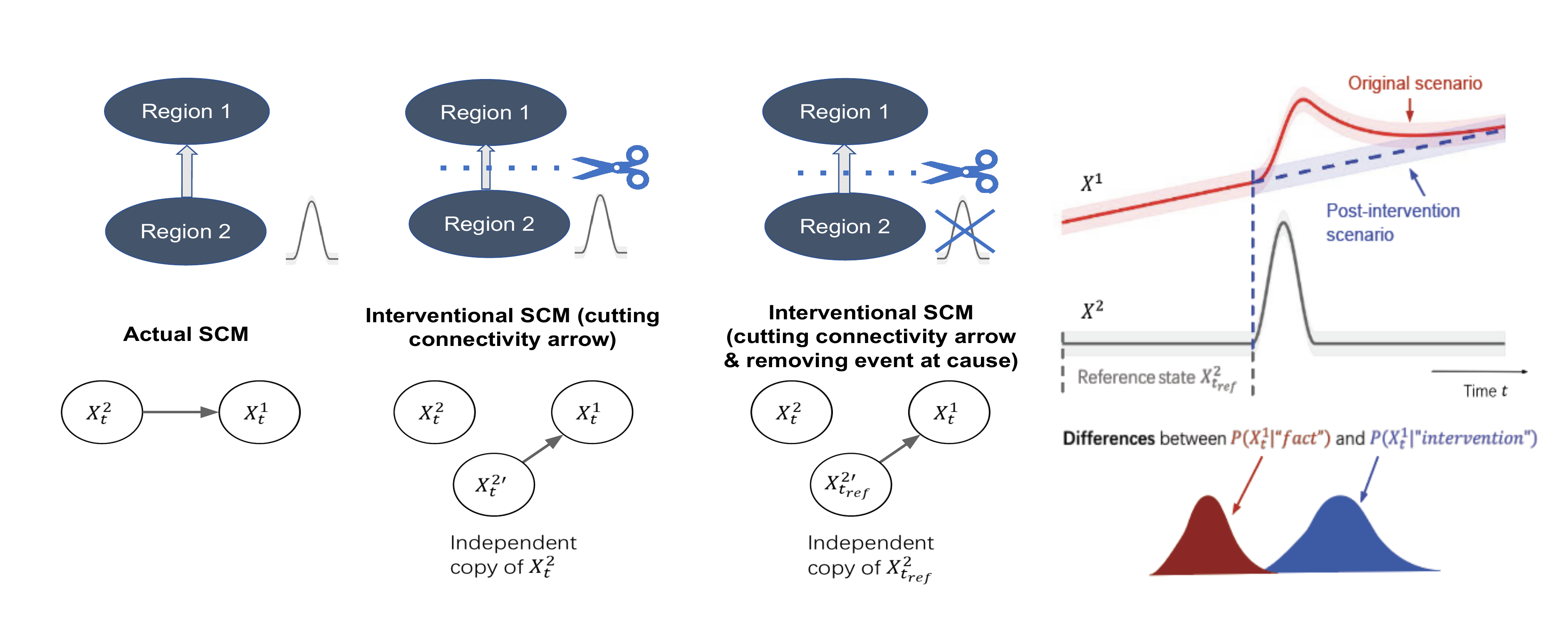}
    \caption{Conceptualizing the intervention for DCS. This diagram illustrates a unidirectional causal link from a cause (Region 2, state $X^2_t$) to an effect (Region 1, state $X^1_t$). To quantify the causal influence within a Structural Causal Model (SCM), the intervention conceptually "cuts" the causal arrow. This is achieved by replacing the input from the cause with an independent copy ($X^{2'}_{p,t}$) drawn from its current marginal distribution, thereby isolating the direct causal effect \cite{shao2023transient}.
    Figure adapted from \citet{shao2023transient}, with modifications.}
    \label{fig:counterfactual}
\end{figure}

\begin{figure}[t]
    \centering
    \includegraphics[width=\textwidth]{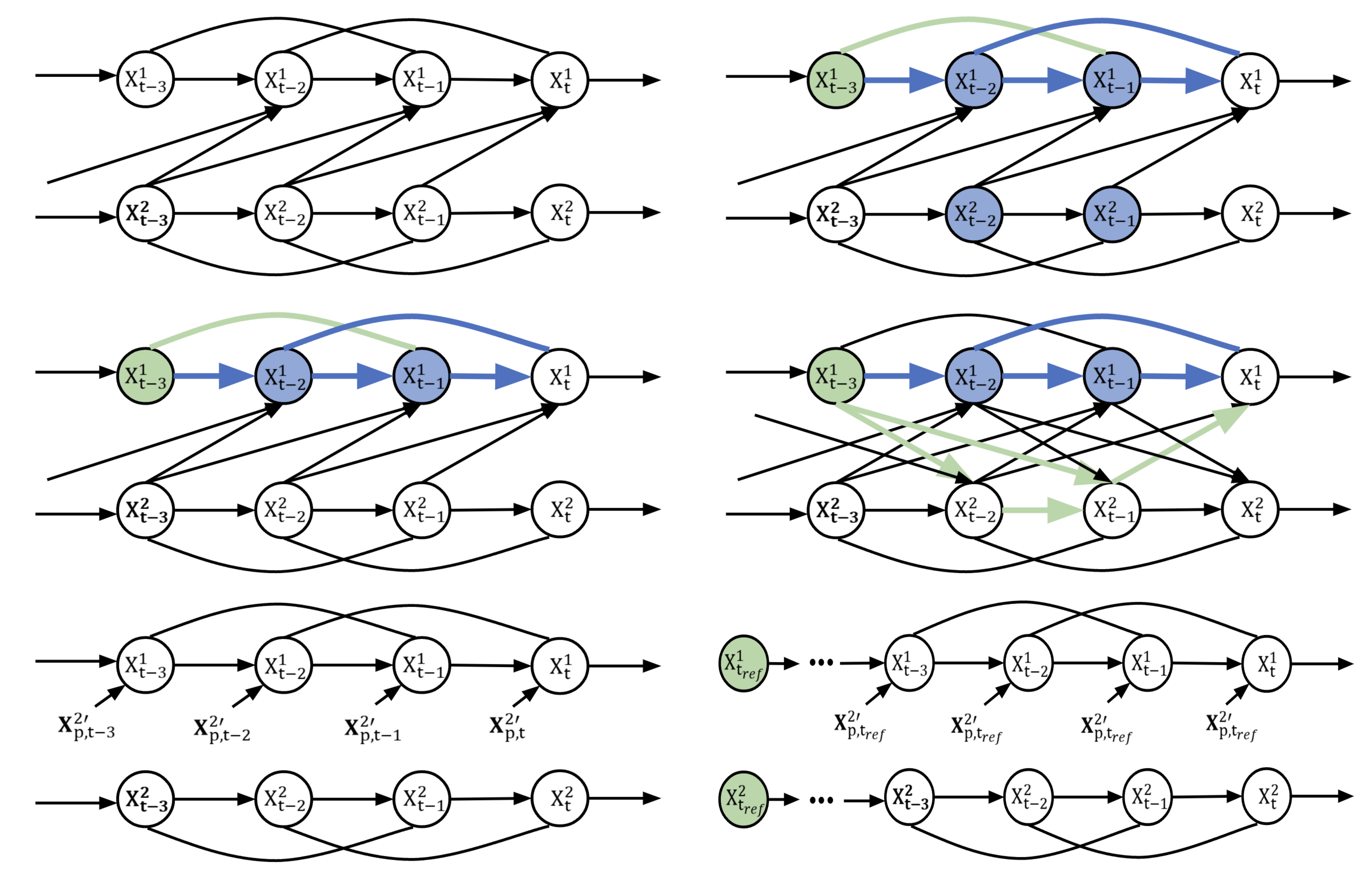}
    \caption{Conceptualizing the intervention for rDCS. To specifically assess event-related causality, the rDCS intervention cuts the causal link $X^2_{p,t} \to X^1_t$ and replaces the input from the cause with a copy ($X^2_{p,t_{ref}}$) drawn from its marginal distribution at a stationary baseline or reference time. This design makes rDCS highly sensitive to event-driven causal interactions \cite{shao2023transient}.
    Figure adapted from \citet{shao2023transient}, with modifications.
    }
    \label{fig:interventsion}
\end{figure}

To conceptualize these interventions, we adapt the diagrams from \citet{shao2023transient} in \autoref{fig:counterfactual} and \autoref{fig:interventsion}. These diagrams visualize the process for peri-event neural data, showing how the causal link $X^2_{p,t} \to X^1_t$ is cut by replacing the influence of $X^2_{p,t}$ on $X^1_t$ with either an independent copy of the cause or a sample from its baseline state.

\noindent \textbf{Intervention (do-operator)} 
Let the bi-variate, time-inhomogeneous SVAR be

\begin{equation}  
    \begin{aligned}
    X^1_t &= a_t^\top X^1_{p,t} + b_t^\top X^2_{p,t} + \eta^1_t, 
    &\quad \eta^1_t \sim \mathcal N(\kappa^1_t, \sigma^2_{1,t}),\\
    X^2_t &= c_t^\top X^1_{p,t} + d_t^\top X^2_{p,t} + \eta^2_t, 
    &\quad \eta^2_t \sim \mathcal N(\kappa^2_t, \sigma^2_{2,t}),
    \end{aligned}
\end{equation}

with $X^k_{p,t} = [X^k_{t-1}, \ldots, X^k_{t-p}]^\top$.

Intervening on $V_k$ replaces its structural assignment 
$V_k := f_k(\mathrm{PA}_k, N_k)$ by 
$V_k := \tilde f_k(\widetilde{\mathrm{PA}}_k, \tilde N_k)$. 
The induced interventional distribution is written $P_{\mathrm{do}(V_k := \tilde f_k)}$ 
and is obtained by combining: 
(i) unchanged conditionals for all \emph{unaffected} assignments, and 
(ii) the new assignment at the intervened node.

\subsection{Two concrete interventional schemes used in this work}

\begin{enumerate}
\item \textbf{''Cut-the-arrow'' (for DCS).} 
Remove $X^2_{p,t}\!\to\!X^1_t$ by feeding $X^1_t$ with an \emph{independent copy} 
$X^{2\prime}_{p,t}\sim p(X^2_{p,t})$ that preserves the \emph{marginal} statistics of the cause but is independent of all exogenous variables. This yields the post-intervention conditional
\begin{equation}
    p_{\mathrm{DCS}}\!\left(X^1_t \mid X^1_{p,t}\right) 
    = \int p\!\left(X^1_t \mid X^1_{p,t}, x^2_{p,t}\right)\, p(x^2_{p,t})\, dx^2_{p,t}.
\end{equation}

DCS is then the expected KL divergence between the original conditional and this post-intervention conditional. This leads to \autoref{eq: dcs} and \autoref{eq: dcs-gaussian-svar}.

\emph{This keeps the cause’s current variability in place while removing the mechanism linking it to the effect.}

\item \textbf{''Cut-the-arrow + baseline cause'' (for rDCS).} 
For peri-event analyses, the cause’s distribution may be deterministically shifted 
(e.g., by stimulus locking). rDCS therefore \emph{replaces} $p(X^2_{p,t})$ above by a \emph{reference (baseline)} marginal $p_{\mathrm{ref}}(X^2_{p,t_{\mathrm{ref}}})$ 
taken from a pre-event period:
\begin{equation}
    p_{\mathrm{rDCS}}\!\left(X^1_t \mid X^1_{p,t}\right) 
    = \int p\!\left(X^1_t \mid X^1_{p,t}, x^2_{p,t}\right)\, 
    p_{\mathrm{ref}}(x^2_{p,t_{\mathrm{ref}}})\, dx^2_{p,t}.
\end{equation}
Therefore, rDCS can be formulated in the form of \autoref{eq: rdcs} and \autoref{eq: rdcs-gaussian-svar}.

\end{enumerate}

\subsubsection{Remarks}

It is essential to highlight that the explicit analytical expressions provided above for TE, DCS, and rDCS are explicitly derived under the assumption of linear Gaussian SVAR models \cite{moneta2011causal}. This assumption simplifies the computation of information-theoretic quantities required by these measures. However, their general definitions, formulated in terms of entropy differences or KL divergences within the SCM framework, offer broader applicability beyond the linear Gaussian context \cite{shao2023transient}.

DCS and rDCS offer notable advantages over traditional measures, such as TE \cite{Schreiber2000, vicente2011transfer} and GC \cite{barnett2009granger}. These advantages primarily arise because DCS and rDCS are grounded in SCMs \cite{janzing2013quantifying, ay2008information}, utilizing hypothetical interventions that explicitly remove causal links to quantify direct causal effects. In contrast, TE and GC rely on Wiener’s predictive principle and measure observational dependencies rather than direct causality. Consequently, the SCM-based approach renders DCS and rDCS inherently local measures, well-suited for analyzing transient, non-stationary events by emphasizing direct causal mechanisms at specific time points \cite{shao2023transient}.

Specifically, DCS addresses the limitations of TE in scenarios involving strongly synchronized signals by avoiding the so-called "vanishing problem," where TE unintuitively decreases despite strong synchronization. Further refining the causal analysis, rDCS excels due to its enhanced sensitivity to deterministic perturbations or shifts in the mean of the innovation’s distribution (a sensitivity that both TE and DCS typically lack). Since event-based neural data commonly feature deterministic components driven by specific events, rDCS employs a baseline reference state in its intervention procedure. This design enables rDCS to effectively capture event-related causal interactions and reveal underlying mechanisms during transient events, making it particularly powerful for analyzing event-driven causality \citep{shao2023transient}.

\paragraph{Alignment \& selection bias.}
When constructing peri-event ensembles, it is important to 
\textbf{align on the putative cause rather than the effect}. 
In the SCM framework, conditioning on detections derived from the cause ensures unbiased estimation of causal influence, whereas aligning on the effect introduces selection bias and can even reverse the apparent 
direction of influence. 
In practice, one can compare both alignments, but the interpretation should be based on cause-aligned results, as illustrated in the neurophysiological example \citep[also referred to as][for details]{shao2023transient}.

\subsection{Simulations}

The simulation module facilitates testing and validation by generating synthetic multivariate AR signals with user-defined parameters:

\begin{itemize}
    \item \textbf{Coupled AR models:} Users can specify coefficients (including zero for no connection) to create ground-truth causal structures (e.g., $X \rightarrow Y$, $Y \rightarrow X$, bidirectional, or none).
    \item \textbf{Parameter control:} Allows customization of model order, noise levels, and time series length.
    \item \textbf{Use cases:} Useful for validating the package's causal inference methods, exploring the impact of parameters like noise or data length on detection power \citep[similar to analyses in ][]{shao2023transient}, and for educational purposes.
\end{itemize}
\section{Example}

\begin{figure}[!tbp]
    \centering
    \includegraphics[width=\textwidth]{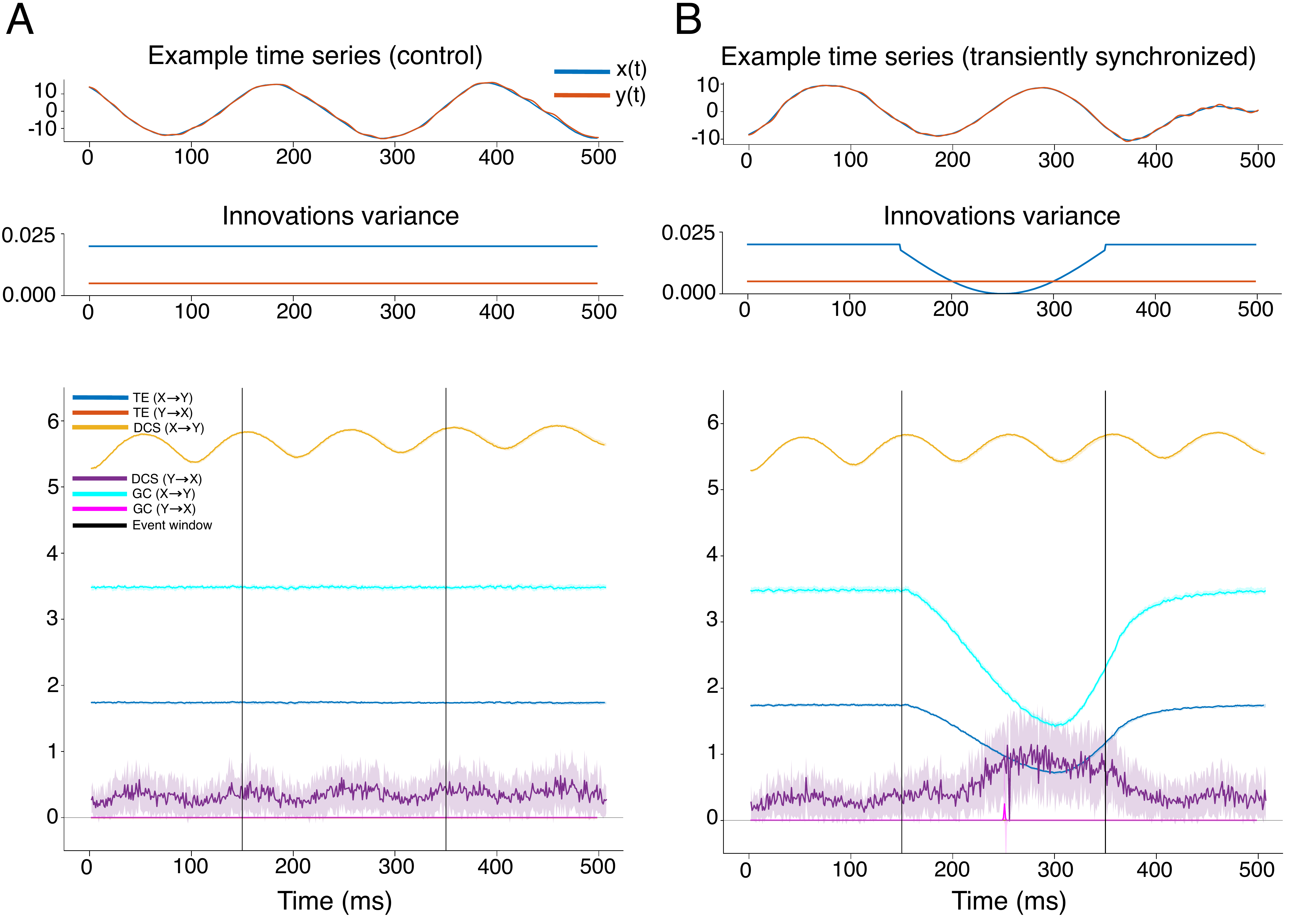}
    \caption{DCS successfully captures causal influence during high synchrony where TE fails, demonstrating a key advantage of the SCM-based approach.
    This figure replicates the 'synchrony pitfall' simulation from \cite{shao2023transient} using the package to illustrate how different causality measures perform under varying levels of signal synchronization. A known causal link exists from X to Y ($X \rightarrow Y$) throughout the simulation.
    \textbf{(A)} Control condition with stationary coupling.
    (top) An example of a coupled time series, where the driver $x(t)$ influences $y(t)$. 
    (middle) The innovation variance for both signals is kept constant, leading to a stable level of synchrony.
    (bottom) In this stable state, both Transfer Entropy (TE, light blue) and Dynamic Causal Strength (DCS, cyan) correctly and consistently identify the directed causal influence from X to Y.
    \textbf{(B)} Transiently synchronized condition.
    (top) The time series becomes visibly more synchronized within the event window (demarcated by dashed lines).
    (middle) This high synchrony is induced by a transient decrease in the innovation variance of the driver signal $x(t)$.
    (bottom) During this event, TE drops significantly toward zero, falsely suggesting that the causal link has weakened or vanished. In contrast, DCS remains high, correctly reflecting the intact underlying causal mechanism. This demonstrates the robustness of DCS for analyzing transient, event-related neural dynamics where synchrony may change.}
    \label{fig:causality}
\end{figure}

\begin{figure}[!tbp]
    \centering
    \includegraphics[width=0.95\textwidth]{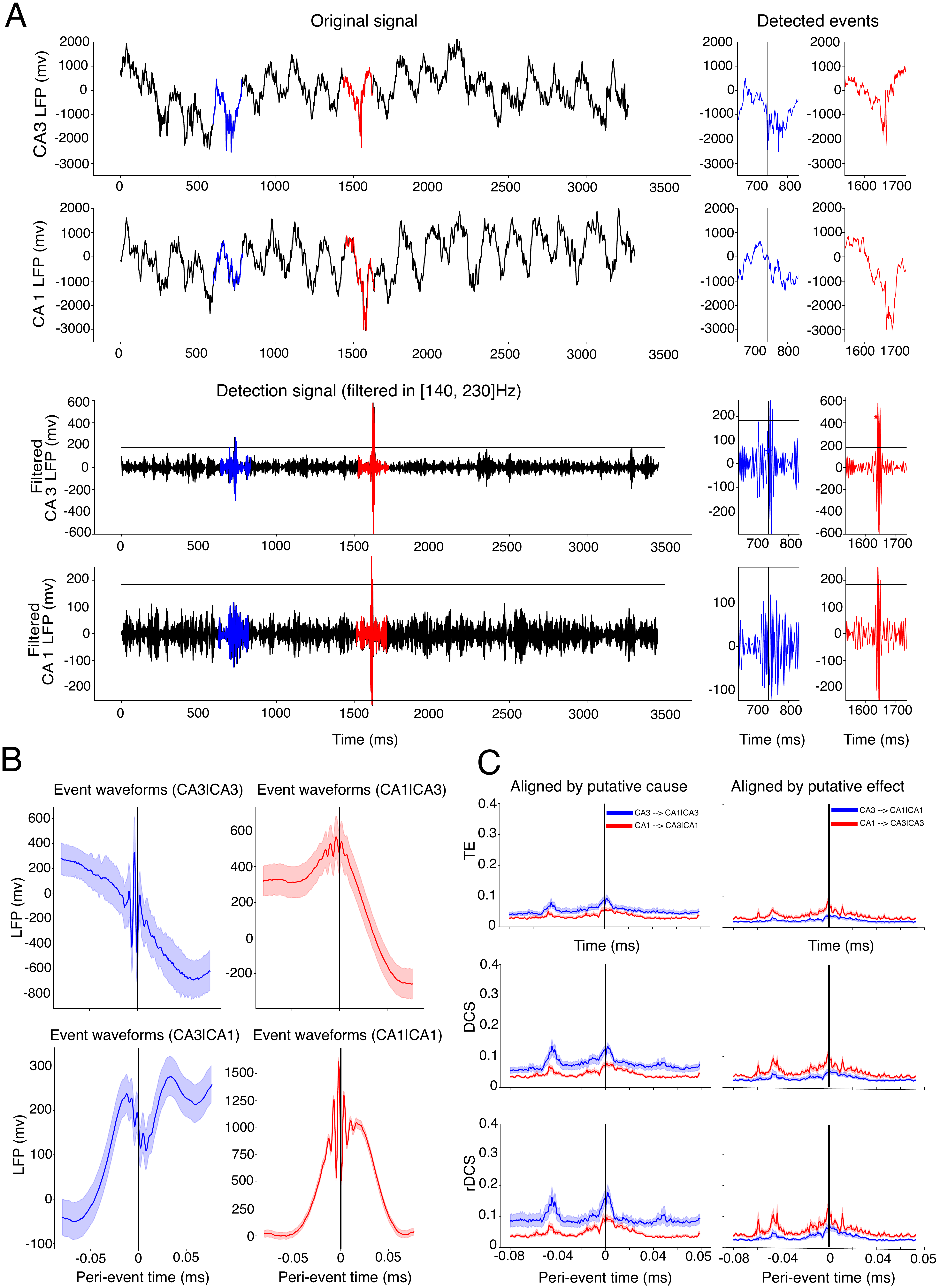}
    \afterpage{
    \captionof{figure}{
    Demonstration of \texttt{trancit} on real-world LFP data, correctly identifying directed causality from hippocampal area CA3 to CA1 during SWR events. This event-based causal analysis highlights the importance of aligning data on the putative cause to obtain accurate results.
    \textbf{(A)} Signal processing and event detection.
    Examples of the original and bandpass-filtered signals from the CA3 and CA1 regions are shown. Two SWR events (marked in blue and red) are highlighted, which were detected by applying a threshold to the putative cause signal (CA3) and aligned by the local peak of the ripple-band activity.
    \textbf{(B)} Event-aligned waveforms.
    The average waveforms of the SWR event ensembles are shown for CA3 and CA1. The top row is aligned by the peak in the CA3 signal, while the bottom row is aligned by the peak in the CA1 signal. Shading represents the standard error across channel pairs.
    \textbf{(C)} Peri-event causality results.
    Causality was measured using TE, DCS, and rDCS for event ensembles aligned by the putative cause (CA3, left column) and the putative effect (CA1, right column). Critically, 
    rDCS (bottom row, left) reveals a significant, transient information flow from CA3 to CA1 only when the analysis is aligned correctly on the cause. 
    When aligned on the effect (right column), this directed influence is obscured. Shading reflects the standard deviation of 100 repeated bootstrap ensembles.
    \rule{\linewidth}{0.5pt}
    \bigskip}
    \label{fig:ca1_ca3_analysis}
}
\end{figure}

\subsection{Causal inference based on synthetic data}

To demonstrate core functionality and validate the implementation, key results from \citet{shao2023transient} were replicated using the \texttt{dcs} package. \autoref{fig:causality} illustrates the application of implemented causality measures (e.g., TE, DCS) to simulated AR signals with a known causal structure ($X \rightarrow Y$), analogous to Figure 4 in the original publication \cite{shao2023transient}.
%
In \autoref{fig:causality}, we replicated the simulated causality detection from \citet{shao2023transient} using the \texttt{dcs} package. The figure illustrates the package's ability to accurately identify directed influence ($X \rightarrow Y$) in synthetic time series that mimic transient neural dynamics.

This example serves as validation and provides a practical guide for users. An accompanying Jupyter notebook detailing this replication is available in \href{https://github.com/CMC-lab/TranCIT/blob/main/examples/dcs_introduction.ipynb}{dcs\_introduction.ipynb} \cite{nouri_2025_trancit}.

Besides validating the toolkit, another critical reason for reproducing \autoref{fig:causality} is to warn users about how to interpret TE around tightly synchronized events, as TE can fail under such scenarios. Specifically, TE can drop to (nearly) zero when the two signals are strongly synchronized --- even if a real causal link exists --- whereas DCS remains informative. 
The reproduced example of \citet{shao2023transient} that demonstrates this with two coupled oscillators: briefly lowering the driver's noise increases synchrony, and \textit{TE transiently decreases}, while DCS does not. Mathematically, in Gaussian SVARs,
\begin{equation}
    \mathrm{TE}(X^2 \to X^1)_t 
    = \tfrac{1}{2}\log 
    \frac{\sigma^2_{1,t} + b_t^\top \operatorname{Cov}(X^2_{p,t} \mid X^1_{p,t}) b_t}
    {\sigma^2_{1,t}}.
\end{equation}
If $X^2$ becomes (almost) a deterministic function of $X^1$ 
(e.g., $X^2_t \approx k X^1_t$), then 
$\operatorname{Cov}(X^2_{p,t} \mid X^1_{p,t}) \to 0$ 
and $\mathrm{TE}\to 0$, 
even though the causal arrow $X^2 \to X^1$ is intact. 
In contrast, DCS depends on the \emph{marginal} covariance of $X^2_{p,t}$,
\begin{equation}
    \mathrm{DCS}(X^2 \to X^1)_t 
    = \tfrac{1}{2}\log 
    \frac{b_t^\top \operatorname{Cov}(X^2_{p,t}) b_t + \sigma^2_{1,t}}
    {\sigma^2_{1,t}},
\end{equation}
so it does \emph{not} vanish when synchrony is high.

This is precisely the behavior shown in the reproduced figure.
In one line, \autoref{fig:causality} reproduces the classic 'synchrony pitfall' for TE --- when two signals march in lockstep, TE can under-report directed influence --- thereby motivating our use of DCS/rDCS for transient events.

\subsection{Application to neurophysiological data}

To demonstrate the practical utility of the TranCIT toolkit on real-world data, we replicated a key analysis of directed causality between the hippocampal CA3 and CA1 subfields using publicly available Local Field Potential (LFP) recordings during sharp wave-ripple events, an analysis inspired by the original methodology paper \citet{shao2023transient}. As illustrated in \autoref{fig:ca1_ca3_analysis}, the package's preprocessing workflow was used to detect transient events and extract aligned data epochs. Subsequently, the rDCS method was applied to quantify the causal influence. The results correctly identify a significant, transient information flow from CA3 to CA1, a well-established pathway in hippocampal memory processing. This example not only validates the \texttt{dcs} implementation against known neuroscientific findings but also serves as a practical blueprint for future applications. The complete code to reproduce this figure is provided as a Python script in \href{https://github.com/CMC-lab/TranCIT/blob/main/examples/lfp_pipeline.py}{lfp\_pipeline.py} \cite{nouri_2025_trancit}. 
We anticipate that a similar analysis will be insightful in several other cases where neural or behavioral transient events may accompany large-scale network interactions \citep[e.g., see][]{rabinovich2008transient,he2018robust,safavi2024decision,spyropoulos2024distinct,vinck2025large,wu2025neural}.

\section{Implementation details}

The \texttt{trancit} package is open-source and distributed under the permissive BSD-2 license, ensuring it can be freely used and incorporated into diverse academic and commercial projects. The package is designed with usability and extensibility in mind:

\begin{itemize}
    \item \textbf{Workflow integration:} Primarily centered around a main analysis pipeline function that takes a structured configuration object, allowing straightforward integration into larger data processing pipelines. This object specifies all parameters for preprocessing (e.g., event detection thresholds, epoch duration) and causality analysis (e.g., model order, choice of measure). This design ensures that analyses are easily configurable, shareable, and reproducible.

    \item \textbf{Robustness:} Includes error handling for invalid configurations and utilizes established numerical libraries (\texttt{NumPy}, \texttt{SciPy}) for core computations.

    \item \textbf{Modularity:} Organized into distinct modules (\texttt{causality}, \texttt{models}, \texttt{simulation}, \texttt{utils}, \texttt{core}, and \texttt{pipeline}) to bring code readability, maintainability, and ease of extension with new methods or features.

    \item \textbf{Testing and community support:} Core functionalities are verified by a suite of unit tests implemented using the \texttt{pytest} framework, ensuring algorithmic reliability. These tests are run automatically on each pull request using GitHub Actions to ensure software reliability. Furthermore, the project welcomes community involvement; guidelines for contributing code, reporting issues via the GitHub repository's issue tracker, and seeking support are provided within the software repository.
\end{itemize}

\section{Acknowledgments}

We acknowledge the foundational work by \citet{shao2023transient} on the dynamic causal strength methodology. We also thank the developers and communities behind the core Python scientific libraries utilized in \texttt{trancit}, including NumPy \cite{harris2020array}, SciPy \cite{virtanen2020fundamental}, and Matplotlib \cite{hunter2007matplotlib}.
KS acknowledges the support from the Shanghai Municipal Science and Technology Major Project (Grant No. 2019SHZDZX02) and the Max Planck Society (including the Max Planck Institute for Biological Cybernetics and the Graduate School of Neural and Behavioral Sciences).
SS acknowledges the add-on fellowship from the Joachim Herz Foundation.

\bibliography{references}

\if@endfloat\clearpage\processdelayedfloats\clearpage\fi 


\end{document}